\documentclass[conference]{IEEEtran}
\IEEEoverridecommandlockouts
\usepackage{cite}
\usepackage{amsmath,amssymb,amsfonts}
\usepackage{algorithmic}
\usepackage{graphicx}
\usepackage{textcomp}
\usepackage{xcolor}
\usepackage{wrapfig}
\def\BibTeX{{\rm B\kern-.05em{\sc i\kern-.025em b}\kern-.08em
    T\kern-.1667em\lower.7ex\hbox{E}\kern-.125emX}}
\begin{document}

\title{ Performance Evaluation of Geospatial Images based on Zarr and Tiff\\
}

\author{
\IEEEauthorblockN{Jaheer khan}
\IEEEauthorblockA{\textit {Center of Data for Public Good} \\
\textit{IISc, Bangalore}\\
Jaheerkhan200201@gmail.com\\
}
\and
\IEEEauthorblockN{ Swarup E}
\IEEEauthorblockA{\textit {Center of Data for Public Good} \\
\textit{IISc, Bangalore}\\
swarup.e1998@gmail.com\\
}
\and
\IEEEauthorblockN{ Rakshit Ramesh}
\IEEEauthorblockA{\textit {Center of Data for Public Good} \\
\textit{IISc, Bangalore}\\
rakshitadmar@gmail.com\\
}

}

\maketitle

\begin{abstract}
This evaluate the performance of geospatial image processing using two distinct data storage formats: Zarr and TIFF. Geospatial images, converted to numerous applications like environmental monitoring, urban planning, and disaster management. Traditional Tagged Image File Format is mostly used because it is simple and compatible but may lack by performance limitations while working on large datasets. Zarr is a new format designed for the cloud systems,that offers scalability and efficient storage with  data chunking and compression techniques. This study compares the two formats in terms of storage efficiency, access speed, and computational performance during typical geospatial processing tasks. Through analysis on a range of geospatial datasets, this provides details about the practical advantages and limitations of each format,helping users to select the appropriate format based on their specific needs and constraints.
\end{abstract}


\section{Introduction}
Geospatial imagery plays a crucial role in various fields like environmental science, agriculture, urban development, and disaster response. With the increasing availability of satellite images and other geospatial data, the need for efficient storage, retrieval, and processing has become more usable. Image formats like TIFF have been  used due to their widespread acceptance and straight forward implementation. The  big data and cloud computing necessitates are required more advanced storage solutions to manage and process large volumes of geospatial data effectively.

Zarr is a relatively new storage format designed for the challenges with big data in geospatial applications. It offers features like chunked storage, which helps for efficient reading and writing of large datasets, and compatible with new cloud storage systems. Unlike TIFF, which stores image data in a continuous manner, Zarr organizes data into manageable chunks, helps for faster access and more efficient parallel processing.

We targeted to provide a evaluation of the performance of geospatial image processing using Zarr and TIFF formats. We used pytorch model and different python libraries to compare these formats on various performance metrics, including storage efficiency, data access speed, and processing time for common geospatial tasks. By analyzing this, we are able to identify the strengths and weaknesses of each format and provide feedback for their use in different scenarios.

The sections will contains the methodology employed in our experiments, present the results of our performance evaluations, and discuss the implimentation of  for the geospatial dataset. This analysis offers valuable insights into the optimal use of Zarr and TIFF formats for geospatial image storage and processing and  contributing to more efficient and effective handling of large-scale geospatial data. 

\section{Litrature Review}
The Pangeo Project enables scientists to handle large datasets using open-source technologies like xarray, dask, and Kubernetes on HPC systems and cloud platforms. This setup allows for scalable data analysis by distributing computations across multiple compute nodes. The main challenge is data delivery speed to these nodes, making data access methods and file formats critical. Traditional netCDF/HDF archives are inefficient for cloud storage as they require reading entire files to extract specific data. Instead, Pangeo employs the zarr format, which separates metadata into a lightweight JSON format and stores data in a chunked, compressed binary format. This allows for independent, thread-safe retrieval of data chunks. This review demonstrates zarr's performance and scalability advantages in a cloud-based Pangeo environment for scientific workflows. [1]

Analyzing large-scale scientific data requires efficient storage and manipulation software. Traditionally, netCDF with HDF5 has been the standard for chunking and compressing large datasets. However, new formats like Zarr are being integrated into netCDF libraries. This study compares read times for netCDF and Zarr in both Java and Python using data from the Community Climate System Model (CCSM). Findings indicate that Zarr reads faster with large chunk sizes, while netCDF-4 with HDF5 becomes more efficient as chunk sizes increase. These differences arise from how each format stores data, affecting read efficiency. Understanding these performance variations is crucial for optimizing data processing in scientific research. [2]

The study evaluated chunking strategies for storing and accessing multi-dimensional Earth science datasets in the cloud by converting data from the Goddard Earth Observing System model from NetCDF to Zarr format. Results showed that chunk size significantly impacted processing time and memory usage, with larger chunks along the target dimension performing best for specific data operations like time series extraction and map generation. Versatile middle-range chunking strategies offered balanced performance, highlighting optimal approaches for cloud-based storage and analysis of large datasets.[3]

Sharing large N-dimensional datasets has been a challenge in various fields, often necessitating substantial server resources to generate cropped or down-sampled representations of terabyte-scale data interactively. Recently, a cloud-native chunked format called Zarr has been adopted in the bioimaging domain to standardize and alleviate this burden. This format holds potential interest for other consortia and sectors within the NFDI. [4]

 The Zarr archive organizes data into manageable two-dimensional files, improving accessibility and reducing memory and processing demands compared to GRIB2. It offers flexibility for accessing specific subdomains and parameters, enhancing efficiency in analyzing high-impact weather events and enabling high-throughput workflows in cloud computing environments. Adoption of Zarr may extend beyond weather science, potentially becoming a standard for data repositories, supported by tools like xarray for diverse research application. [5]

Xarray is a Python package and open-source project designed for N-dimensional labeled arrays, integrating an API inspired by pandas with the Common Data Model for self-describing scientific data. It features label-based indexing and arithmetic, compatibility with core Python scientific libraries (such as pandas, NumPy, and Matplotlib), support for out-of-core computation on large datasets, extensive serialization and input/output options, and advanced tools for multi-dimensional data manipulation like group-by and resampling. Widely embraced in geoscience, xarray also finds applications in diverse fields such as physics, machine learning, and finance for comprehensive multi-dimensional data analysis. [6]

In this paper, Authers outlined foundational concepts for processing .tiff images, exploring diverse options for their manipulation. Our approach is versatile and can be applied as a comprehensive framework for various GIS analysis tasks. This paper serves as a valuable resource for individuals engaged in image processing, offering insights and methodologies applicable across different applications in the field [7]

\section{USE CASES}

1. Rasterio is a Python library for working with geospatial raster data. It have various use cases such as: 

a) Rasterio can read various raster formats such as GeoTIFF and extract metadata and data arrays. It can write data arrays to new raster files.

b) It is better to perform operations on the small dataset( Tiff) images but for the large datasets it is slower then the zarr.

c) Retrieve and update metadata, such as coordinate reference systems and band description.

d) Change the projection of raster data to match different CRS. Resample rasters to different spatial resolutions or grid alignments.

e) Efficiently read raster datasets in smaller chunks or windows to manage memory usage and process large files incrementally.

f) Generate reduced-resolution versions of raster datasets for faster access and visualization\\

2. Zarr is a format for the store the chunked, compressed, N-dimensional array. Appending all chunks to a Zarr array can have use cases such as:

  a) Continuously append data from sensors or IoT devices as new data comes in.Collect and append data from ongoing experiments without needing to reload the entire dataset.

  b) It is better to perform operations on the large dataset( Tiff) images but for the smaller datasets it is slower then the rasterio.
 
  c) Append intermediate results of data processing steps to a Zarr array, enabling checkpointing and fault tolerance.

  d) Store and append large satellite images or other geospatial data, which are typically too large to handle in-memory.

  e) Store and append genomic sequencing data, which often come in large volumes and need efficient storage.

  f) Store large datasets used for training machine learning models in chunks, allowing for incremental updates as new data is collected.

\section{METHODOLOGY}
 We used the methods to  evaluate the performance of the geospatial images:\\
 1. Using Xarray Library\\
 2. Using Zarr Library\\
 3. Using Rasterio Libarary\\

 1. Using Xarray Library-  Array is an open source project and Python package that represents labels in the form of dimensions and attributes on NumPy-like arrays. Xarray includes a large and growing library of  functions for advanced operations and visualization.
 
  Geospatial data obtained as .tiff images using Python library Xarray, to handle the multi-dimensional nature of the data. Initially, we read the .tiff files using xarray. We normalized the  values to a range suitable for neural network input, between 0 and 1. Then integrating this data with PyTorch by converting the Xarray into PyTorch tensors. We implemented a custom PyTorch Dataset class to manage the conversion process and included necessary data. This class was then used with PyTorch's DataLoader to handles batching, shuffling, and parallel data loading.

 Performance evaluation was carried out using metrics such as CPU/GPU usage and memory uses. we analyzed the model’s predictions to understand its performance and identify potential improvements. This comprehensive methodology ensured an effective application of PyTorch on .tiff images using xarray, from data preprocessing to model evaluation.\\

 2. Using Zarr Library - Zarr is a Python package providing an implementation of compressed, chunked, N-dimensional arrays, designed for use in parallel computing. we targeted to process geospatial data stored in .tiff images using Zarr and PyTorch to evaluate performance .The .tiff files were read using the Rasterio library, which is well to handling geospatial data. Then the data  converted to the Zarr format, which supports chunked, compressed, N-dimensional arrays.

 The data stored in Zarr format was normalized to a range suitable for neural network input, between 0 and 1. We implemented a custom data loading pipeline that  converts the Zarr array into PyTorch tensor. A custom PyTorch Dataset class was developed for loading of data in batches and any necessary data. Using PyTorch DataLoader, we are batching, shuffling, and parallel data loading. 

Performance evaluation was carried out using metrics such as CPU/GPU usage and memory uses. we analyzed the model’s predictions to understand its performance and identify potential improvements. This comprehensive methodology ensured an effective application of PyTorch on .tiff images using Zarr, from data preprocessing to model evaluation.

3. Using Rasterio Library -  Rasterio is a library that is  working with geospatial raster data in Python. It used the capabilities of the  providing a more Pythonic interface for reading geospatial datasets. Rasterio supports a variety of raster file formats, including GeoTIFF, which is commonly used in remote sensors . With Rasterio, users can perform a wide range of operations on raster data such as reading specific bands, extracting metadata, applying transformations. The library also allows for efficient reading and writing of large datasets through windowed operations, enabling users to handle large geospatial datasets without loading the entire dataset into memory. Rasterio works well with other Python libraries such as NumPy and Xarray.

\section{IMPLEMENTATION}

   We need to evaluate the performance of reading the data using PyTorch.we must assess the performance of various operations performed on the data.\\

   For evalute the reading the data performance we use three methods -:

   1. Using the Tiff and xarray -

   i) Data Loading and Preprocessing-: We implemented a custom PyTorch Dataset class, to load geospatial data from a TIFF file using Xarray. This class reads the TIFF file with in Xarray, extracting data for each band. Data normalization implemented using aNormalize class to standardize values by subtracting the mean and dividing by the standard deviation.

 ii) Resource Usage Measurement-: We use utility function, measure\_resource\_usage, to quantify the resource consumption during data loading. This function measures elapsed time, CPU usage, and memory usage using the psutil and time modules. We applied this function to monitor the performance of our data loading function, load\_data, which iterates over a DataLoader configured for efficient batch loading without additional worker processes.

 iii). Performance Evaluation-: The results from measure\_resource\_usage provided crucial insights into the efficiency of our data loading and preprocessing pipeline. This evaluation included the time taken for loading batches, CPU utilization, and memory footprint, ensuring that our approach is optimized for handling large geospatial datasets effectively in machine learning workflows.\\





 2. Using zarr without appending chunks-

 i) Data Loading and Preprocessing-: We developed a custom PyTorch Dataset class, Dataset, which reads data from a Zarr file using Xarray. The data is accessed through the specified variable 'band\_data', and a normalization transformation is applied to standardize the pixel values. This dataset is then used to create a DataLoader that handles batching and shuffling of the data, with a batch size of 4  for data loading.

 ii) Resource Usage Measurement-: To evaluate the efficiency of our data loading process, we implemented a measure\_resource\_usage function. This function measures the elapsed time, CPU usage, and memory consumption of the data loading operation. The function was applied to the load\_data function, which iterates over the DataLoader, printing each batch of data to verify successful loading and processing.

 iii) Performance Evaluation-: The results from the measure\_resource\_usage function provided valuable insights into the performance of our data loading pipeline. We observed the time taken, CPU utilization, and memory usage during the data loading process.\\

 3. Using Zarr with appending all chunks in a zarr array-

 i) Extracting Valid Data Windows from Multi-dimensional Array-:  Iterates over a multi-dimensional array test\_ar to extract overlapping data windows. It checks each window against specific criteria: ensuring minimal zeros in the 9th band (chip\_mask\_size) and sufficient positive values in the 1st band (chip\_cartosat\_size). Valid windows meeting these conditions are appended to zarr\_array\_final, with counters tracking the total processed windows and valid appends.

 ii) Data Loading and Preprocessing-: We implemented a PyTorch Dataset class, ZarrDataset, designed to load data from a Zarr array efficiently. The class utilizes the Zarr array provided during initialization and defines methods (\_\_len\_\_ and \_\_getitem\_\_) to facilitate batch loading of data. Each data sample is converted into a PyTorch tensor of type torch.float32 for compatibility with neural network models.

 iii) Resource Usage Measurement-: To assess the resource efficiency of our data loading process, we introduced the measure\_resource\_usage function. This function utilizes the psutil and time modules to measure the elapsed time, CPU usage, and memory consumption during the execution of a specified function (load\_data). This function was applied to monitor the performance of our DataLoader, configured with a batch size of 4 and no additional worker processes .

 iv) Performance Evaluation-: The evaluation results from measure\_resource\_usage provided essential metrics for assessing the performance of our data loading pipeline. We recorded the elapsed time, CPU utilization, and memory usage during the execution of the DataLoader.\\

 For the performing the operation we used two methods -:\\
 So we performed the mean operation for evaluating the perfromance on the operations. There are two methods we used:\\

  1). Rasterio -

   i)Measure Resource Usage -: We developed the measure|\_resource\_usage function to evaluate performance by measuring elapsed time, CPU usage, and memory consumption during function execution using the psutil library.
   
   ii) Calculate Mean The calculate\_mean function computes the mean of a given data array using NumPy's np.mean.

   iii) Process Zarr Chunks-: The process\_zarr\_chunks function processes data chunks from a Zarr array. It iterates over the chunks, calculating the mean, elapsed time, CPU usage, and memory usage for each chunk using measure\_resource\_usage. It accumulates these metrics to provide an overall performance summary.

The Zarr array is processed with process\_zarr\_chunks, resulting in mean values for all chunks, total elapsed time, total memory usage, total CPU usage, and the average mean value.\\

 2) Zarr-

  i) Measure Resource Usage-: We defined measure\_resource\_usage to measure elapsed time, CPU usage, and memory usage during the execution of a specified function using psutil and time modules.

ii) Calculate Mean-: The calculate\_mean function computes the mean of a given data array using NumPy's np.mean.

iii) Process Zarr Chunks-: In process\_zarr\_chunks, we iterate through chunks of a Zarr array, calculating the mean, elapsed time, CPU usage, and memory usage for each chunk using measure\_resource\_usage. We accumulate these metrics to provide an overall performance summary.

The Zarr array loaded\_zarr is processed with process\_zarr\_chunks, yielding mean values for all chunks, total elapsed time, total memory usage, total CPU usage, and the average mean value. These metrics are printed to summarize performance and resource consumption.

\section{Results}

Results for the comparing the reading time of the dataset using the three different methods where Method-1 - Using the Tiff and xarray,Method-2 -Using zarr without appending chunks,Method-3 -Using Zarr with appending all chunks in a zarr array.
\begin{figure}[h!] 
  \centering
  \includegraphics[width=0.45\textwidth]{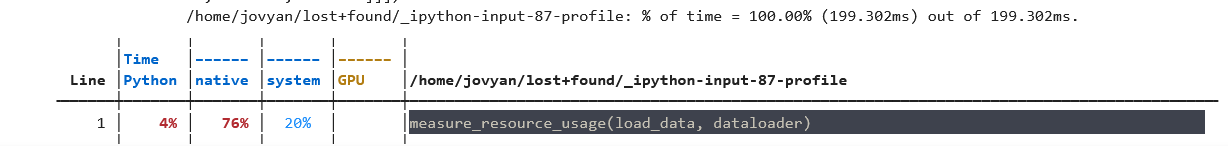} 
  \caption{Using the Tiff and xarray}
  \label{fig:universe} 
\end{figure}


\begin{figure}[h!] 
  \centering
  \includegraphics[width=0.45\textwidth]{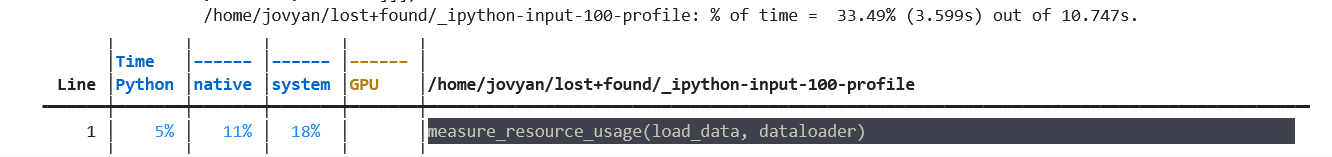} 
  \caption{Using zarr without appending chunks}
  \label{fig:universe} 
\end{figure}

\begin{figure}[h!] 
  \centering
  \includegraphics[width=0.45\textwidth]{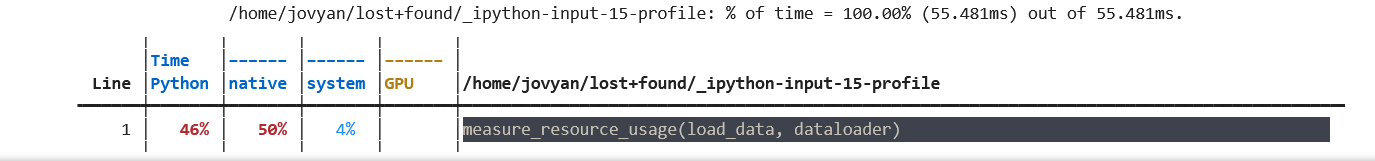} 
  \caption{Using Zarr with appending all chunks in a zarr array}
  \label{fig:universe} 
\end{figure}






\renewcommand{\arraystretch}{2}
\begin{table}[h!]
\centering
\caption{Comparisions of reading the data using pytorch}
\label{places}
\begin{tabular}{|c|c|c|c|c|}
\hline
Method No & Python Time(\%) & Native(\%) & System(\%)  & Time(s)\\
\hline
Method-1 &   4  &  76    &  20  &0.199302 \\
\hline
Method-2 &  5   &  11     & 18   &10.747 \\
\hline
Method-3 &  46   &  50     & 4   &  0.055481\\
\hline
 
\end{tabular}
\renewcommand{\arraystretch}{4.5}
\vspace{1mm}
\end{table}

\begin{figure}[h!] 
  \centering
  \includegraphics[width=0.45\textwidth]{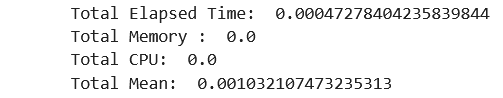} 
  \caption{Using rasterio for performing mean operation}
  \label{fig:universe} 
\end{figure}

\begin{figure}[h!] 
  \centering
  \includegraphics[width=0.45\textwidth]{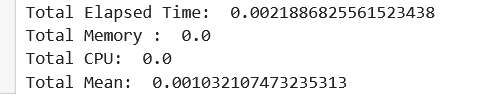} 
  \caption{Using Zarr for performing mean operation}
  \label{fig:universe} 
\end{figure}

we also evaluate the performance for the performing the mean operation on the tiff dataset using the rasterio and zarr library.
\renewcommand{\arraystretch}{2}
\begin{table}[h!]
\centering
\caption{Comparisions of performing mean operation}
\label{places}
\begin{tabular}{|c|c|c|c|c|}
\hline
Libray & Total Eclapsed Time& Mean \\
\hline
Rasterio & 0.00047278404235839844 &  0.001032107473235313    \\
\hline
Zarr &  0.0021886825561523438  &   0.001032107473235313    \\
\hline

\end{tabular}
\renewcommand{\arraystretch}{4.5}
\vspace{1mm}
\end{table}

As we can see that rasterio is performing better than the zarr.\\
\section{CONCLUSION AND FUTURE WORK}

 Future work can improve the methodologies implemented by integrating advanced python libraries and improving model robustness. Using distributed computing frameworks such as Dask or Apache Spark can helps to processing of large geospatial datasets. Developing a dynamic resource management system will optimize computational resources based on data loading requirements. Real-time monitoring and visualization of resource usage can provide deeper information and fine-tuning of data pipelines. Expanding support for various geospatial data formats and emerging standards will helps to make it scalable and adaptable in diverse applications.

 This Document integrated geospatial data processing with machine learning frameworks, by converting TIFF files to Xarray datasets and zarr and implementing various data loading strategies with PyTorch. Custom PyTorch Dataset classes and data loaders effectively handled data from Xarray, Rasterio, and Zarr. Resource usage measurement functions provided data for optimizing the data loading pipeline. The  analysis of different methods highlighted their strengths and limitations, guiding future improvements. This work establishes a robust framework for processing large-scale geospatial data in machine learning workflows.

\vspace{12pt}

\end{document}